\documentclass[11pt,a4paper]{article}
\usepackage[hyperref]{acl2021}
\usepackage{times}
\usepackage{latexsym}
\usepackage{graphicx}  

\usepackage{subfigure}
\usepackage{multirow}

\usepackage{amssymb}
\usepackage{pifont}
\newcommand{\cmark}{\ding{51}}%
\newcommand{\xmark}{\ding{55}}%

\usepackage{microtype}

\usepackage{xcolor}

\aclfinalcopy 
\def\aclpaperid{1808} 

\title{Are Missing Links Predictable? An Inferential Benchmark for \\ Knowledge Graph Completion}

\author{Yixin Cao$^1$$^2$ \quad Xiang Ji$^1$ \quad Xin Lv$^3$\\
\textbf{Juanzi Li$^3$ \quad Yonggang Wen$^1$ \quad Hanwang Zhang$^1$}\\
$^1$School of CSE, Nanyang Technological University, Singapore\\
$^2$S-Lab, Nanyang Technological University, Singapore\\
$^3$Department of CST, Tsinghua University, Beijing, China\\
{\tt caoyixin2011@gmail.com, lv-x18@mails.tsinghua.edu.cn} \\
{\tt lijuanzi@tsinghua.edu.cn, \{ygwen,hanwangzhang\}@ntu.edu.sg}\\
}

\date{}

\begin{document}
\maketitle
\begin{abstract}
  We present InferWiki, a Knowledge Graph Completion (KGC) dataset that improves upon existing benchmarks in inferential ability, assumptions, and patterns. First, each testing sample is predictable with supportive data in the training set. To ensure it, we propose to utilize rule-guided train/test generation, instead of conventional random split. Second, InferWiki initiates the evaluation following the open-world assumption and improves the inferential difficulty of the closed-world assumption, by providing manually annotated negative and unknown triples. Third, we include various inference patterns (e.g., reasoning path length and types) for comprehensive evaluation. In experiments, we curate two settings of InferWiki varying in sizes and structures, and apply the construction process on CoDEx as comparative datasets. The results and empirical analyses demonstrate the necessity and high-quality of InferWiki. Nevertheless, the performance gap among various inferential assumptions and patterns presents the difficulty and inspires future research direction. Our datasets can be found in \url{https://github.com/TaoMiner/inferwiki}.
  
  
\end{abstract}

\section{Introduction}

Knowledge Graph Completion (KGC) aims to predict missing links in KG by inferring new knowledge from existing ones. Attributed to its reasoning ability, KGC models are crucial in alleviating the KG's incompleteness issue and benefiting many downstream applications, such as recommendation~\cite{cao2019unifying} and information extraction~\cite{hu2021how,cao2020learning}. However, the KGC performance on existing benchmarks are still unsatisfactory --- 0.51 Hit Ratio@1 and 187 Mean Rank of the top-ranked model~\cite{wang2019knowledge} on the widely used FB15k237~\cite{toutanova2015observed}. Do we have a slow progress of models~\cite{akrami2020realistic}? Or should we blame for the low-quality of benchmarks?

\begin{table}[tbp]
  \small
  \centering
  \begin{tabular}{|r|c|c|c|} \hline
    & \textbf{Head} & \textbf{Predicate} & \textbf{Tail} \\ \hline
    \textbf{Test} & David & location & ? (Ans: Florida) \\ \hline
    \textbf{Train} & David & placeOfBirth & Atlanta  \\
     & David & nationality & U.S.A.  \\ \hline \hline
     \textbf{Test} & Zurich & travelMonth & ? (Ans: October) \\ \hline
     \textbf{Train} & Zurich & travelMonth & Jan., Feb., Mar., Apr.,  \\
       & & & May., Jun., Jul., Aug.,\\ 
       & & & Sep., Nov., Dec. \\ \hline
  \end{tabular}%
\caption{Low-quality examples in FB15k237. We only present related triples. Ans denotes the missing entity.}
\label{tab:example}
\end{table}%

In this paper, we re-think the task of KGC and construct a new benchmark dubbed InferWiki that highlights three fundamental objectives:

\textbf{Test triples should be inferential}: this is the essential requirement of KGC. Each test triple should have supportive samples in the train set. However, we observe two major issues of current KGC datasets: unpredictable and meaningless test triples, which may hinder evaluating and advancing state-of-the-arts. As shown in Table~\ref{tab:example}, the first example of inferring the location for David (i.e., Florida) is even impossible for humans --- not to mention machines --- merely based on his birthplace and nationality (i.e., Atlanta and USA). In contrast, the second one is predictable but meaningless to find the missing month from a list of months within a year. The above cases are very common in existing datasets, e.g., YAGO3-10~\cite{dettmers2018convolutional} and CoDEx~\cite{safavi2020codex}, mainly due to their construction process: first collecting a high-frequency subset of entities and then randomly splitting their triples into train/test. In this setting, KGC models may be over- or under-estimated, as we are even unsure if a human can perform better.

\begin{table*}[]
  \centering
  \small
  \begin{tabular}{|p{1.3cm}|c|c|c|c|c|c||c|c|}
  \hline
   & \textbf{FB15k237} & \textbf{WN18RR} & \textbf{YAGO3-10} & \textbf{CoDEx-m} & \textbf{Kinship} & \textbf{Country} & \multicolumn{2}{c|}{\textbf{InferWiki16k/64k}} \\ \hline
   \textbf{Source} & FreeBase & WordNet & YAGO & Wikidata & \multicolumn{2}{c||}{Artificial} & \multicolumn{2}{c|}{Wikidata} \\ \hline
  \textbf{Inferential} & \xmark & \xmark & \xmark & \xmark & \cmark & \cmark & \cmark & \cmark \\ \hline
  \textbf{\#Entity} & 14,541 & 40,943 & 123,182 & 17,050 & 104 & 272 & 16,288 & 64,718 \\ \hline
  \textbf{\#Relation} & 237 & 11 & 37 & 51 & 26 & 2 & 197 & 239 \\ \hline
  \textbf{\#train} & 272,115 & 86,835 & 1,079,040 & 185,584 & 8,548 & 1,111 & 162,424 & 782,243 \\ \hline
  \textbf{\#valid} (+) & 17,535 & 3,034 & 5,000 & 10,310 & 1,069 & 24 & 3,398 & 7,747 \\
  (-$\backslash$UNK) & -$\backslash$- & -$\backslash$- & -$\backslash$- & 10,310$\backslash$- & -$\backslash$- & -$\backslash$- & 1,910$\backslash$1,456 & 6,125$\backslash$1,605 \\ \hline
  \textbf{\#test} (+) & 20,466 & 3,134 & 5,000 & 10,311 & 1,069 & 24 & 3,398 & 7,747 \\ 
  (-$\backslash$UNK) & -$\backslash$- & -$\backslash$- & -$\backslash$- & 10,311$\backslash$- & -$\backslash$- & -$\backslash$- & 1,868$\backslash$1,501 & 6,062$\backslash$1,685 \\ \hline
  \end{tabular}
  \caption{Statistics of KGC datasets. A more detailed survey table can be found in Appendix~\ref{sec:apd_rl}.}
  \label{tab:stat}
\end{table*}

\textbf{Test triples may be inferred positive, negative, or unknown}. Following open-world assumption: what is not observed in KG is not necessarily false, but unknown~\cite{shi2018open}. However, existing benchmarks generate unseen triples as negatives (i.e., the closed-world assumption), because KG contains only positive triples. They usually randomly corrupt the head or tail entity in a triple, sometimes with type constraints~\cite{li2019semi}. This leads to trivial evaluation (almost 100\% accuracy in triple classification~\cite{safavi2020codex}). Besides, the lack of unknown test ignores a critical inference capacity and may cause false negative errors in knowledge-driven tasks~\cite{kotnis2017analysis}.

\textbf{Inference has various patterns}. Concentrating on limited patterns in evaluation may bring in severe bias. Domain-specific datasets Kinship~\cite{kemp2006learning} and Country~\cite{bouchard2015approximate} only focus on a few relations and are nearly solved~\cite{das2017go}. General-domain WN18RR~\cite{dettmers2018convolutional} contains prevalent symmetry relation types, which incorrectly boosts the performance of RotatE~\cite{abboud2020boxe}. Clearly, limited patterns leads to unfair comparisons among KGC models.

To this end, we curated an \textbf{Infer}ential KGC dataset extracted from \textbf{Wiki}data and establish the benchmark with two settings of varying in sizes and structures: \textbf{InferWiki64k} and \textbf{InferWiki16k}.
Instead of random split, we mine rules via AnyBURL~\cite{AnyBURL} to guide train/test generation. All test triples are thus guaranteed inferential from training data. To avoid the rule leakage, we utilize two sets of triples: a large set for high-quality rule extraction and a small set for train/test split. Moreover, we infer unseen triples and manually annotate them with positive, negative and unknown labels to improve the difficulty of evaluation following both closed-world and open-world assumptions. For inference patterns, we include and balance triples with different reasoning path length, relation types and patterns (e.g., symmetry and composition).

Our contributions can be summarized as follows:

\begin{itemize}
  \item We summarize three principles of KGC: inferential ability, assumptions and patterns, and construct a rule-guided dataset.
  \item We highlight the importance of negatives and unknowns, and initiate open-world evaluation.
  \item We conduct extensive experiments to establish the benchmark. The results and deep analyses verify the necessity and challenge of InferWiki, providing insights for future research.
\end{itemize}

\section{Related Work}

We can roughly classify current KGC datasets into two groups: inferential and non-inferential datasets. The first group is usually manually curated to ensure each testing sample can be inferred from training data through reasoning paths, while they only focus on specific relations, such as Families~\cite{garcia2015composing}, Kinship~\cite{kemp2006learning}, and Country~\cite{bouchard2015approximate}. The limited scale and inference patterns make them not challenging. HOLE~\cite{nickel2016holographic} achieves 99.7\% ACU-PR on the dataset of Country.

The second group of datasets are automatically derived from public KGs and randomly split positive triples into train/test, leading to a risk of testing samples non-inferential from training data. Popular datasets include FB15k-237~\cite{toutanova2015observed}, WN18RR~\cite{dettmers2018convolutional}, and YAGO3-10~\cite{dettmers2018convolutional}. CoDEx~\cite{safavi2020codex} argues the scope and difficulty of the above datasets, thus propose a comprehensive dataset with manually verified hard negatives.

In fact, inference is an important ability for intelligence. Various fields study how inference is done in practice, ranging from logic to cognitive psychology. Inference helps people make reliable predictions, which is also an expected ability for AI models. Indeed, once deployed, a model may have to make a prediction when there is no evidence in the training set. But, instead of an unreliable guess, we highlight the ability to know unknown, a.k.a. open-world assumption.
Therefore, we aim to curate an large-scale inferential benchmark InferWiki including various inference patterns and testing samples (i.e., positive, negative, and unknown), for better evaluation. We list the statistics in Table~\ref{tab:stat}.

\section{Dataset Design}

We describe our dataset construction that comprises four steps: data preprocessing, rule mining,  rule-guided train/test generation, and inferred test labeling. We then give a detailed analysis.

\subsection{Data Preprocessing}
More and more studies utilize Wikidata\footnote{\url{https://www.wikidata.org/}} as a knowledge resource due to its high quality and large quantity. We utilize the September 2019 English dump in experiments. Data preprocessing aims to define relation vocabulary and extract two sets of triples from Wikidata: a large one for rule mining $\mathcal{T}^r$ and a relatively small one for dataset generation $\mathcal{T}^d$. The reason for using two sets is to avoid the leakage of rules. In other words, some frequent rules on the large set may be very few on the small set. The different distributions shall avoid that rule mining methods will easily achieve high performance. Besides, more triples can improve the quality of mined rules. In contrast, the relatively small set is enough for efficient KGC training and evaluation.

In specific, we first extract all triples that consist of two entity items and one relation with English labels. We then remove the repeated triples and obtain 40,199,175 triples with 7,734,841 entities and 1,170 different relation types. Considering rule mining efficiency, we reduce the relation vocabulary by (1) manually filtering out meaningless relations, such as movie ID or film rating, (2) removing relations of \textit{InstanceOf} and \textit{subClassOf} following existing benchmarks~\cite{toutanova2015observed}, (3) select the most frequent 500 relation types. We focus on the most frequent 800,000 entities, which result in 8,632,777 triples as the large set for rule mining. To obtain the small set for dataset construction, we further select the most frequent 120,000 entities and 300 relations, which result in 1,283,246 triples. Note that we also infer new triples and label them as positive, negative, or unknown later.

\subsection{Rule Mining}
Since developing advanced rule mining models is not the focus of this paper and several mature tools are available online, such as AMIE+~\cite{galarraga2015fast} and AnyBURL~\cite{AnyBURL}. We utilize AnyBURL\footnote{\url{http://web.informatik.uni-mannheim.de/AnyBURL/}} in experiments due to its efficiency and effectiveness.

Given a set of triples (i.e., the large set $\mathcal{T}^r$), this step aims to automatically learn rules $\mathcal{F}=\{(f_p,\lambda_p)\}_{p=1}^P$, where $f_p$ denotes a horn rule, e.g., \textit{spouse}$(x,y)$ $\wedge$ \textit{father}$(x,z)$ $\Rightarrow$ \textit{mother}$(y,z)$, and $\lambda_p \in [0,1]$ denotes the confidence of $f_p$. For each rule $f_p$, the left side of $\Rightarrow$ is called the premise, and the right side is called the conclusion, where the conclusion contains a single atom and the premise is a conjunction of several atoms in the Horn rule scheme. We can ground specific entities to replace $x,y,z$ in $f_p$, which shall denote an inferential relationship between premise and conclusion triples. For example, given \textit{spouse}(LeBron James, Savannah Brinson) and \textit{father}(LeBron James, Bronny James), we may infer a new triple \textit{mother}(Savannah Brinson, Bronny James).

Of course, not all of the mined rules are reasonable. To alleviate the negative impacts of unreasonable rules, we rely on more data (a large set of triples) and keep high-confidence rules only. Particularly, we follow the suggested configuration of AnyBURL. We run it for $500$ seconds to ensure that all triples can be traversed at least once and obtain 251,317 rules, where 168,996 out of them whose confidence meets $\lambda_p>0.1$ have been selected as the rule set to guide dataset construction.

\subsection{Rule-guided Dataset Construction}
\label{sec:rgen_data}

Different from existing benchmarks, InferWiki provides inferential testing triples with supportive data in the training set. Moreover, it aims to include as many inference patterns as possible and these patterns are better evenly distributed to avoid biased evaluation. Thus, this step has four objectives: rule-guided split, path extension, negative supplement, and inference pattern balance.

\noindent\textbf{Rule-guided Split} grounds the mined rules $\mathcal{F}$ on triples $\mathcal{T}^d$ to obtain premise triples and corresponding conclusion triples. All premise triples form a training set, and all conclusion triples form a test set. Thus, they are naturally guaranteed to be inferential. For correctness, all of premise triples must exist in the given triple set $\mathcal{T}^d$, while conclusion triples are not necessarily in $\mathcal{T}^d$ and may be generated for further annotation (i.e., Section~\ref{sec:labeling}).

For example, given a rule \textit{spouse}$(x,y)$ $\wedge$ \textit{father}$(x,z)$ $\Rightarrow$ \textit{mother}$(y,z)$, we traverse all of the given triples and find entities \textit{LeBron James}, \textit{Savannah Brinson}, and \textit{Bronny James} that meet the premise. We then add the premise triples \textit{spouse}(LeBron James, Savannah Brinson) and \textit{father}(LeBron James, Bronny James) into the training set, and generate the conclusion triple \textit{mother}(Savannah Brinson, Bronny James) for testing, no matter it is given or not.

\noindent\textbf{Path Extension} aims to increase the inference path patterns by (1) adding more reasoning paths for the same testing triple, and (2) elongating paths by replacing those premise triples that have reasoning paths. For example, we replace \textit{father}(LeBron James, Bronny James) with two triples that can infer it: \textit{father}(LeBron James, Bryce James) and \textit{brother}(Bronny James, Bryce James). The original path is then extended by one hop. Correspondingly, we define the confidence of extended paths as the multiplication of all involved rules. Longer paths will challenge long-distance reasoning ability.


\noindent\textbf{Negative Supplement} is to generate negative triples if we cannot annotate the same number of negatives with positive triples. Otherwise, we will face an imbalance issue.
Following conventions, we randomly corrupt the head or tail entities in a positive triple with the following constraints: (1) the relation of the positive triple is exclusive, e.g., \textit{placeOfBirth}, if the ratio from head to tail entities is smaller than a threshold (we choose $1.2$ heuristically in experiments); otherwise, the corrupted negative triple may be actually positive, leading to false negative errors. (2) We choose positive triples from the test set for corruption to improve the difficulty --- the model has to correctly infer the corresponding positive triple from training data, then classify the corrupted triple as negative through the confliction.
Particularly, for non-exclusive relation types, most of their corrupted results should be unknown following open-world assumption. The inferred test set covers such cases, which will be discussed in Section~\ref{sec:labeling}.

\noindent\textbf{Inference Pattern Balance} aims to balance various inference patterns, including path length, relation types, and relation patterns (i.e., symmetry, inversion, hierarchy, composition, and others). This is because concentrating on some patterns may lead to severe bias and unfair comparison between KGC models~\cite{zhang2020few}. We first count the frequency of testing triples according to the path lengths, relation types and patterns, respectively. For each of them, we rank their counting and choose highest ranked groups of triples as frequent ones, instead of setting a threshold. We then carefully remove some frequent triples randomly, until the new distributions reach an accepted range (checked by humans).

\subsection{Inferred Test Triple Labeling}
\label{sec:labeling}

Different from existing datasets, InferWiki aims to include positive, negative, and unknown testing triples, to evaluate the model under two types of assumptions: open-world assumption and closed-world assumption. The main difference between them is whether unknown triples are regarded as negatives. That is, the open-world evaluation is a three-class classification problem (i.e., positive, negative, and unknown). The closed-world evaluation targets only positive and negative triples, and we can simply relabel unknown triples as negatives without changing the test set.

\begin{table*}[]
  \small
  \centering
  \begin{tabular}{l|p{14cm}}
  \hline
  Test (+)  & Japan National Route 1$_{Q1191191}$, \textit{connectsWith}, Japan National Route 4$_{Q1055023}$ \\ \hline
  Train & Japan National Route 4, \textit{terminus}, Japan National Route 1 \\ \hline\hline
  Test (+)   &   Agnese$_{Q2726556}$, \textit{sibling}, Lucia$_{Q3838490}$ \\ \hline
  Train &  Maddalena$_{Q329555}$,	\textit{sibling},	Agnese $\wedge$ Maddalena, \textit{mother}, Beatrice$_{Q51089}$ $\wedge$ Valentina, \textit{mother}, Beatrice$\wedge$ Viridis$_{Q271827}$, \textit{sibling}, Valentina$_{Q943180}$ $\wedge$ Viridis, \textit{sibling}, Estorre$_{Q3733572}$ $\wedge$ Elisabetta$_{Q1941886}$, \textit{sibling}, Estorre $\wedge$ Lucia, \textit{sibling}, Elisabetta  \\ \hline\hline
  Test (-)  &  Quimper$_{Q702161}$,	\textit{capital},	Versailles$_{Q621}$ ($\perp$ Yvelines$_{Q12820}$,	\textit{capital},	Versailles)  \\\hline
  Train &  Yvelines, \textit{replaces}, Seine-et-Oise$_{Q979470}$ $\wedge$ Seine-et-Oise, \textit{capital}, Versailles   \\ \hline\hline
  Test (-)  &  Roberto$_{Q53003}$, \textit{placeOfBirth}, Baton$_{Q28218}$ ($\perp$ Roberto, \textit{placeOfBirth}, Rome$_{Q220}$) \\\hline
  Train &  Renzo$_{Q1397252}$, \textit{sibling}, Roberto $\wedge$ Renzo, \textit{placeOfBirth}, Rome \\ \hline\hline
  Test (UNK)  &  Midōsuji Line$_{Q1192413}$,	\textit{connectsWith},	Keiyō Line$_{Q741145}$  \\\hline
  Train &  Shin-Ōsaka Station$_{Q801438}$,	\textit{connectingLine},	Midōsuji Line $\wedge$ Tōkaidō Shinkansen$_{Q660895}$,	\textit{terminus},	Shin-Ōsaka Station $\wedge$ Tōkaidō Shinkansen, \textit{connectsWith}, Keihin-Tōhoku Line$_{Q1197028}$ $\wedge$ Keiyō Line, \textit{connectsWith}, Keihin-Tōhoku Line\\ \hline\hline
  Test (UNK)   &  Mary$_{Q104109}$,	\textit{workLocation},	London$_{Q84}$   \\\hline
  Train &  Mary, \textit{memberOfPoliticalParty}, Republican Party$_{Q29468}$ $\wedge$ Carl$_{Q127437}$, \textit{memberOfPoliticalParty}, Republican Party $\wedge$ Carl$_{Q127437}$ \textit{workLocation},	London $\wedge$ Mary, \textit{occupation}, actor$_{Q33999}$ $\wedge$ Mary, \textit{spouse}	Owen$_{Q966972}$ \\ \hline
  \end{tabular}
  \caption{Positive, negative, and unknown examples of InferWiki, where the triples with brackets are not in train set (inferred from related training triples), $\perp$ denotes contradicted triples and subscripts denote Wikidata ID. }
\label{tab:example_inferwiki}
\end{table*}

So far, we have two test sets: one is generated via rule guidance, and the other contains the supplemented negatives. This section aims to label the generated triples. First, we automatically label the triples with positive if they exist in Wikidata. Then, we manually annotate the remaining 4,053 triples. The annotation guideline can be found in Appendix~\ref{sec:apd_ann}. Note that all of the unknowns are factually incorrect but not inferential. To assess the quality of annotations, we verify a random selection of $300$ test triples ($100$ for each label). The annotators agree with our labels $84.3$\% of the time. We further investigate the disagreements by relabeling $100$ samples. 85\% of the time, humans prefer an unknown, while automatic labeling tends to assign them with positive or negative labels. This suggests the inferential difference between humans and machines --- the capacity of knowing unknown.

Finally, we remove the entities that are not in any of the grounded paths and their triples. We randomly select half of the test set as valid. This forms \textbf{InferWiki64k}. We further extract a dense subset \textbf{InferWiki16k} by filtering out the positive triples whose confidence is smaller than $0.6$. Correspondingly, negative/unknown triples are reduced to keep balance. The statistics is listed in Table~\ref{tab:stat}.

\subsection{Dataset Analysis}
\label{sec:data_ana}

Table~\ref{tab:example_inferwiki} shows positive, negative, and unknown examples of InferWiki and their (possible) supportive training data. For positives, their paths seem reasonable and vary in length, relation types, and patterns. The 7-hop path of the sibling example is even difficult for a human. For negatives and unknowns, they are indeed incorrect and more challenging. There are no direct contradicted triples in the train set --- the model is encouraged to reason related triples and justify if there is a confliction (i.e., negative) or not (i.e., unknown). Nevertheless, there are two minor issues. First, some unreasonable paths may corrupt the predictability. We thus increase the rule confidence threshold $\lambda>0.6$ for InferWiki16k and manually annotate uncertain test triples for the correctness of labels. More advanced rule mining models can improve the construction pipeline. We leave it in the future. Second, does unknown triples have a bias on certain relation types? The answer is yes but not exactly. As shown in Table~\ref{tab:example_inferwiki}, the relation $connectsWith$ is involved in both positive and unknown triples, which is also determined by the paths.

Next, we analyze the relation patterns and path length distribution through comparisons with existing KGC datasets. Due to the different construction pipelines, existing datasets are difficult to offer quantitative statistics. We thus apply our pipeline on CoDEx~\cite{safavi2020codex}. Only inferential test triples remain, and the training set keeps unchanged, namely CoDEx-m-infer, which reduces the test and valid positives from 20,622 to 7,050. This agree with the original paper that reports 20.56\% triples are symmetry or compositional through AMIE+ analysis. We find more paths due to more extensive rules extracted from a large set of triples. This also demonstrates the necessity of rule-guided train/test generation --- most test triples are not guaranteed inferential when using random split. 

\begin{figure}[tb]
  \centerline{\includegraphics[width=0.45\textwidth]{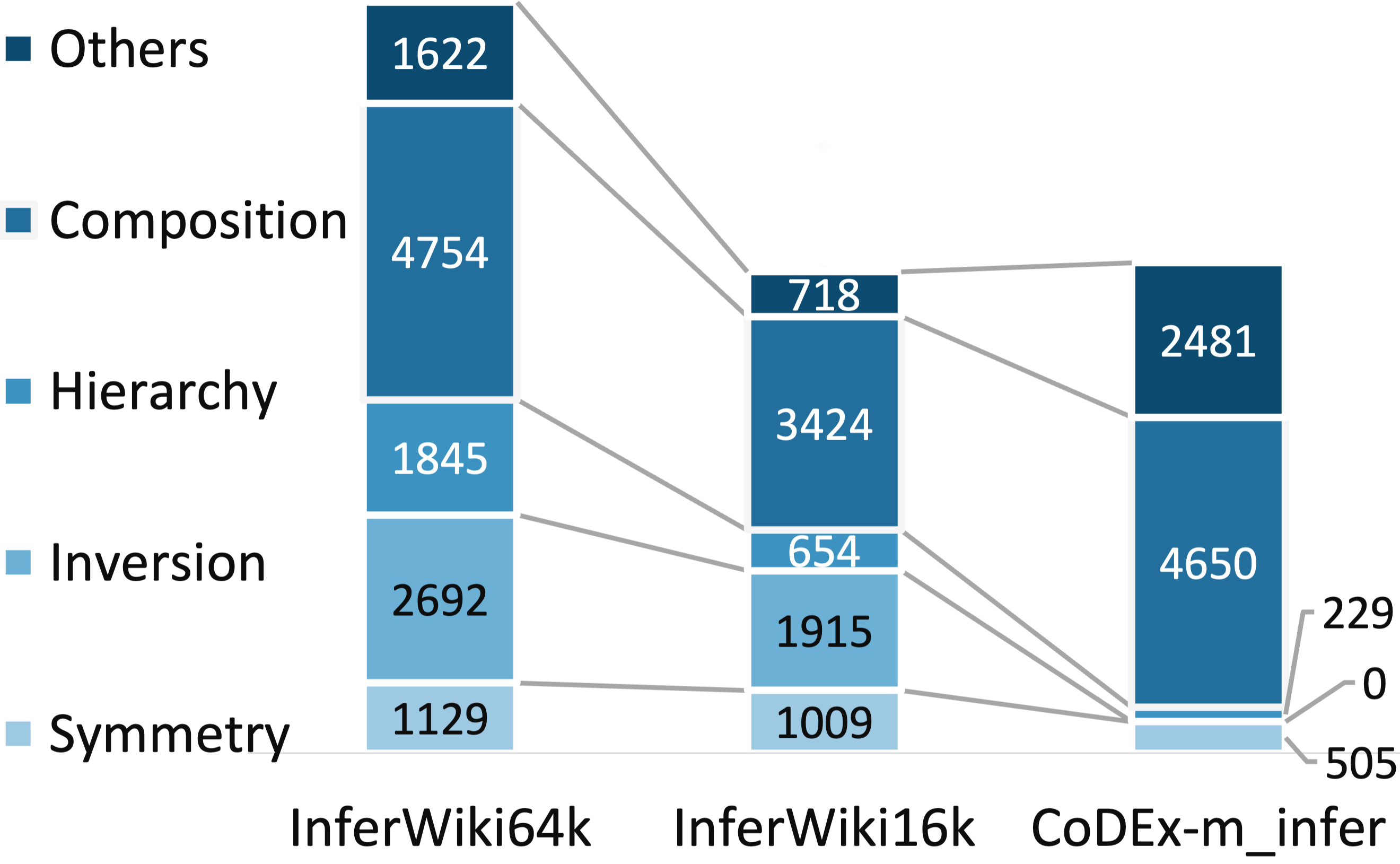}}
    \caption{Distribution of paths in relation patterns.}
  \label{fig:comp_rp}
  \vspace{-0.3cm}
\end{figure}

\noindent\textbf{Relation Pattern} Following convention, we count reasoning paths for various patterns: symmetry, inversion, hierarchy, composition, and others, whose detailed explanations and examples can be found in Appendix~\ref{sec:apd_rp}. If a triple has multiple paths, we count all of them. As Figure~\ref{fig:comp_rp} shows, we can see that (1) there are no inversion and only a few symmetry and hierarchy patterns in CoDEx-m, as most current datasets remove them to avoid train/test leakage. But, we argue that learning and remembering such patterns are also an essential capacity of inference. It just needs to control their numbers for a fair comparison. (2) The patterns of InferWiki is more evenly distributed. Note that the patterns of symmetry, inversion, and hierarchy refer to 1-hop paths, while composition and others refer to multi-hop paths. So, the total number of the former three is almost the same as that of the latter two, to balance paths with varying lengths, which will be discussed next.

\begin{table*}[]
  \small
  \centering
  \begin{tabular}{r|cccc|cccc|cccc}
  \hline
  \multicolumn{1}{l|}{} & \multicolumn{4}{c|}{\textbf{InferWiki64k}} & \multicolumn{4}{c|}{\textbf{InferWiki16k}} & \multicolumn{4}{c}{\textbf{CoDEx-m-infer}} \\ \cline{2-13} 
  \multicolumn{1}{l|}{} & Acc     & Prec    & Recall    & F1    & Acc    & Prec    & Recall    &   F1 & Acc    & Prec    & Recall    &   F1  \\ \hline
  \textbf{TransE}   &  .823  & .782   & .895 &  .835 &  .796 & .736 &  \textbf{.926} & .820 & .763 & .792 & .891 & .839 \\
  \textbf{ComplEx}  & .812 & .779 & .872 & .823 & .811 & .835 & .778  & .805 & .798 & .805 & .936 & .866   \\
  \textbf{RotatE}     & .852  &  .808 & \textbf{.924} &  .862 & .811  & .769 &  .891 & .825 & .788 & .790 & .945 & .861  \\
  \textbf{ConvE}     & \textbf{.881} &  .864 & .906 & \textbf{.884} & \textbf{.897} & \textbf{.887} & .911 & \textbf{.899} & \textbf{.851} & .853 & \textbf{.948} & \textbf{.898}  \\
  \textbf{TuckER}      & .862 & \textbf{.897} & .817 & .855 & .861 & .836 & 899 & .866 & .803 & \textbf{.919} & .784 & .846 \\ \hline
  \end{tabular}
  \caption{Overall Performance of Triple Classification (Closed-world Assumption), where acc and prec stand for accuracy and precision, respectively. }
\label{tab:perf_tc}
\end{table*}

\begin{figure}[tb]
    \centerline{\includegraphics[width=0.45\textwidth]{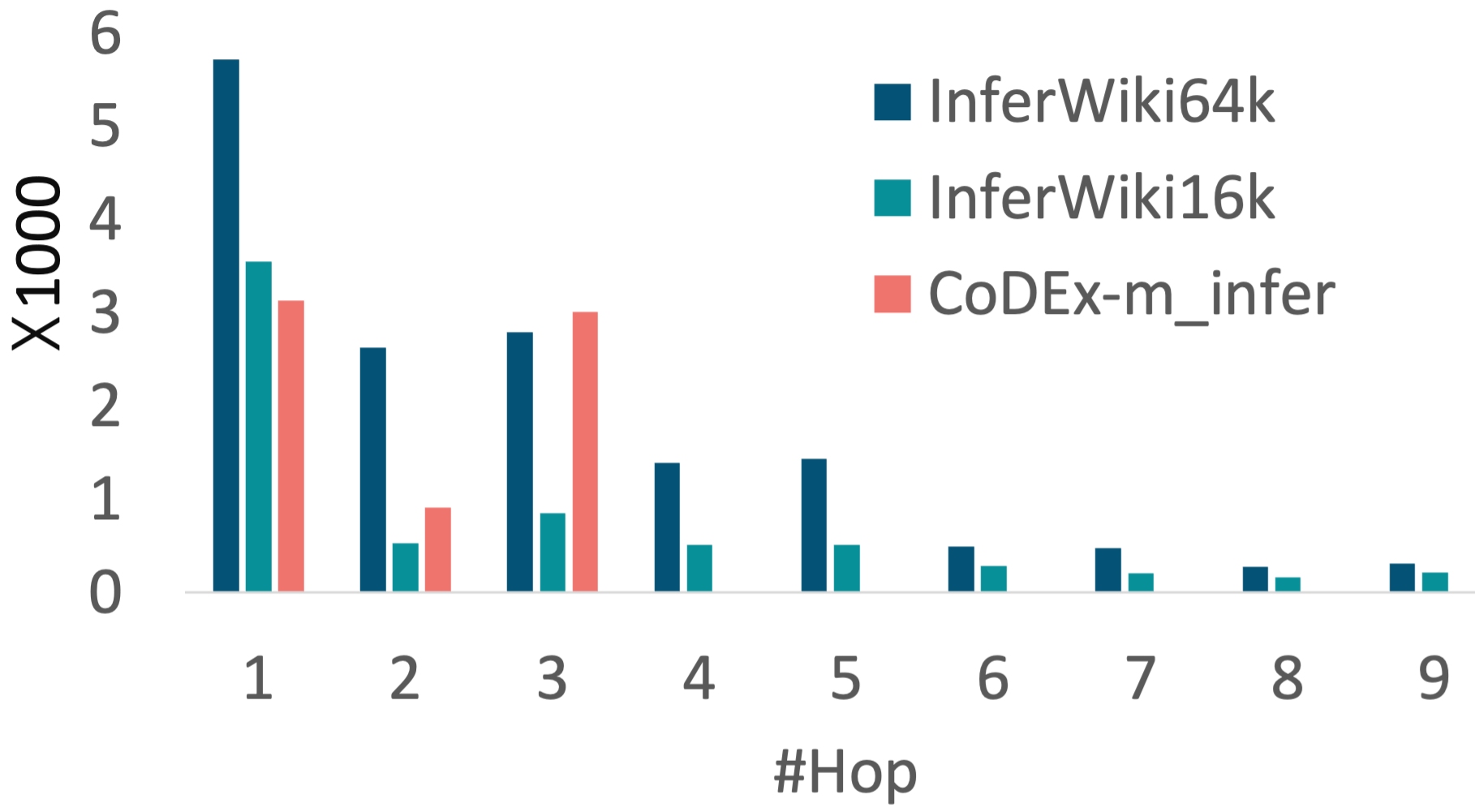}}
      \caption{Comparison of paths in different lengths.}
    \label{fig:comp_pl}
    \vspace{-0.3cm}
\end{figure}

\noindent\textbf{Path Length Distribution} The reasoning paths can ensure test triples' predictability but may not be the shortest ones, as there may be undiscovered paths connecting two entities. Thus, our statistics concerning path length offer a conservative analysis and give an upper bound. For a test triple with multiple paths, we count the shortest one. As shown in Figure~\ref{fig:comp_pl}, we can see that InferWiki has more long-distance paths, while CoDEx-m-infer normally concentrates on maximum 3-hop reasoning paths. In specific, the maximum path length of InferWiki is $9$ (4 before path extension) and the average length is $2.9$ ($1.5$ before path extension).

Further analysis of relation, entity and neighbor distributions can be found in Appendix~\ref{sec:apd_rt}\&\ref{sec:apd_comp}.

\subsection{Limitation}
\label{sec:limitation}
Although we carefully design the construction of inferWiki, there are still two types of limitations: rule biases and dataset errors, that can to be addressed along with the development of KG techniques in the future. In terms of rule biases, AnyBURL may be over-estimated due to its role in the construction. Although we utilize two triple sets to avoid rule leakage, their overlap may still bring unfair performance gain to AnyBURL. We consider synthesize several rule mining results to improve InferWiki in the next version. In terms of dataset errors, first, to balance positive and negative triples in the larger InferWiki64k, we follow conventions to randomly sample a portion of negatives. These negatives may be unknown if following open-world assumption. We manually assess the randomly sampled negatives and find a 15.7\% error rate. Therefore, we conduct open-world experiments on the smaller InferWiki16k, all of whose testing negatives are verified by humans. The second type of errors is due to unreasonable rules for dataset split, which is caused by prediction errors of existing rule mining models. However, there is no suitable evaluation in this field to provide quantitative analysis. Our ongoing work aims to develop an automatic evaluation for path rationality to improve the mining quality, and thus facilitate our inferential pipeline.

\section{Benchmarking}

\subsection{Tasks}
We benchmark performance on InferWiki for the tasks: (1) \textbf{Link Prediction}, the task of predicting the missing head/tail entity for a given query triple (?, r, t) or (h, r, ?). Models are encouraged to rank correct entities higher than others in the vocabulary. We adopt the filtering setting~\cite{bordes2013translating} that excludes those entities, if the predicted triples have been seen in the train set. Mean reciprocal rank (MRR) and hits@k are standard metrics for evaluation. (2) \textbf{Triple Classification} aims to predict a label for each given triple (h, r, t). The label following open-world assumption is trinary $y\in \{-1, 0, 1\}$ and becomes binary $y\in \{-1, 1\}$ when adopting closed-world assumption --- all $0$-label triples are re-labeled with $-1$, since our unknown triples are factually negative yet non-inferential from training data. Since KGC models output real-value scores for triples, we classify scores into labels by choosing one or two thresholds per relation type on valid. Accuracy, precision, recall, and F1 are measurements.

\subsection{Models}

For comprehensive comparison, we choose three types of representative models as baselines: (1) Knowledge Graph Embedding models, including \textbf{TransE}~\cite{bordes2013translating}, \textbf{ComplEx}~\cite{trouillon2016complex}, \textbf{RotatE}~\cite{sun2019rotate}, \textbf{ConvE}~\cite{dettmers2018convolutional}, and \textbf{TuckER}~\cite{balazevic2019tucker}, (2) multihop reasoning model \textbf{Multihop}~\cite{MultiHop}, and (3) rule-based \textbf{AnyBURL}~\cite{AnyBURL}. Note that the latter two are specially designed for link prediction. The detailed implementation including parameters and thresholds can be found in Appendix~\ref{sec:apd_exp}.

\subsection{Triple Classification Results}

Table~\ref{tab:perf_tc} shows micro scores for triple classification. We can see that all of the baselines perform well --- around 90\% F1 scores. This is consistent with recent findings that triple classification is a nearly solved task (around 98\% F1 scores)~\cite{safavi2020codex}. Nevertheless, the lower performance demonstrates the difficulty of our curated datasets, mainly due to the manually annotated hard negatives of InferWiki (and CoDEx).

\noindent\textbf{Impacts of Hard Negatives}

\begin{figure}[tb]
  \centerline{\includegraphics[width=0.47\textwidth]{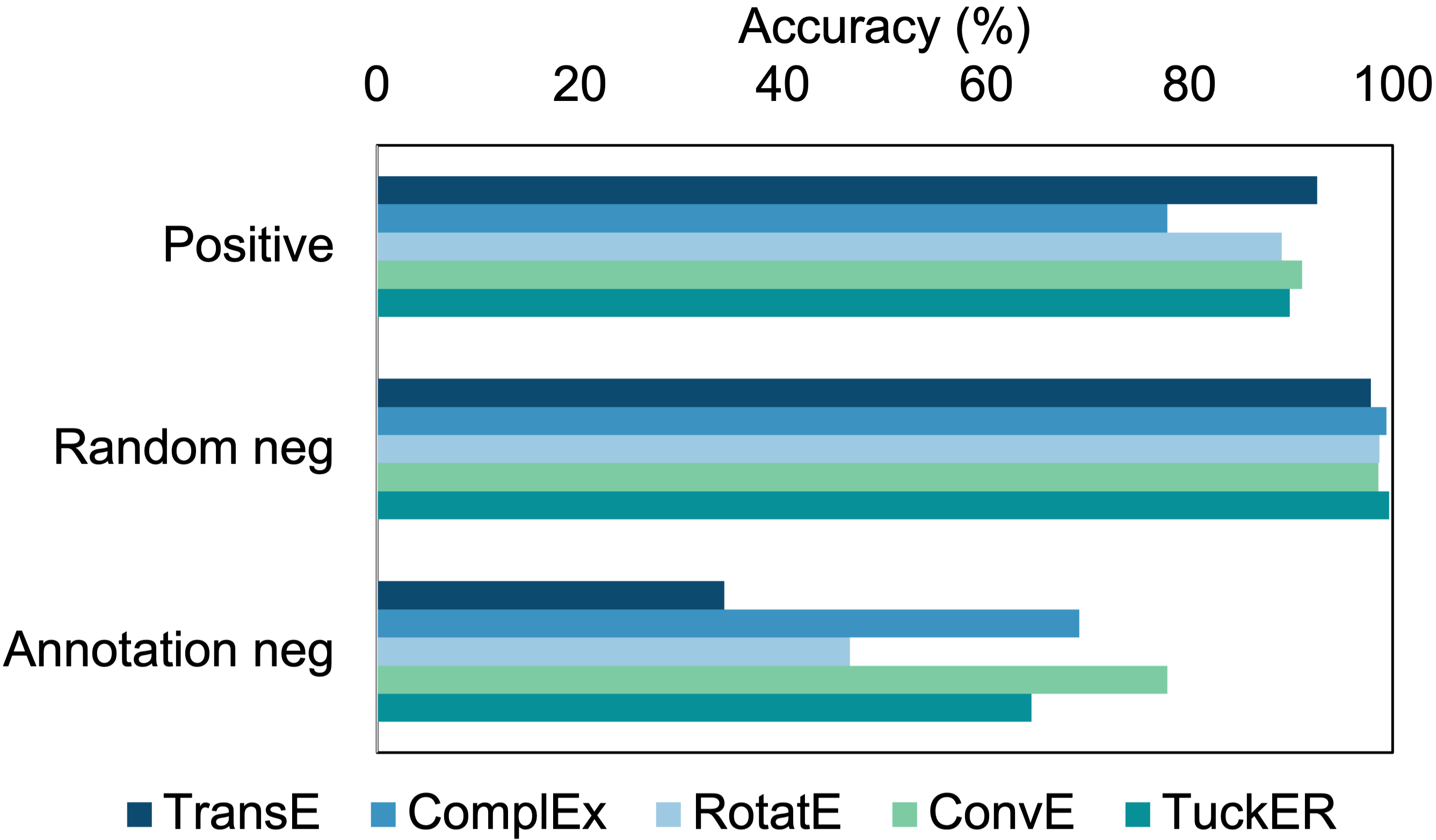}}
    \caption{Accuracy regarding various negative types, where random neg denotes supplemented negatives, and annotation neg denotes annotated negatives (including unknowns).}
  \label{fig:neg}
\end{figure}

Figure~\ref{fig:neg} presents the accuracy on InferWiki16k regarding various types of triples: positive, random supplemented negatives, and annotated negatives (including relabeled unknowns). We can see that (1) random negative triples are indeed trivial for all of baseline models, which motivates the necessity of harder negative triples to push this research direction forward, (2) positive triples are slightly difficult to judge than random negatives, and (3) the accuracy significantly drops on annotation negatives. This is mainly because most annotated triples are actually unknown --- they are factually incorrect, but there are no obvious abnormal patterns. Such non-inferential cases may underestimate KGC models.

\noindent\textbf{Open-world Assumption}

\begin{table}[t]
  \small
  \centering
  \begin{tabular}{r|c|c|c|c}
  \hline
  \multicolumn{1}{l|}{} & Acc & Prec &  Recall & F1 \\ \hline
  \textbf{TransE}       &  .711  &  .687   &  .668  & .676  \\
  \textbf{ComplEx}      & .723  & .703 &  .709 & .701  \\
  \textbf{RotatE}       & .745  & .746 &  .750 & .736  \\
  \textbf{ConvE}        & \textbf{.803} & \textbf{.763} & \textbf{.777} &  \textbf{.768}  \\
  \textbf{TuckER}       & .709  & .639 &  .657  &  .618	 \\ \hline
  \end{tabular}
  \caption{Performance of Triple Classification on InferWiki16k (Open-world Assumption).}
\label{tab:perf_open_tc}
\end{table}

\begin{table*}[]
  \small
  \centering
  \begin{tabular}{|r|ccc|ccc|ccc|}
  \hline
  \multicolumn{1}{|l|}{} & \multicolumn{3}{c|}{\textbf{InferWiki64k}} & \multicolumn{3}{c|}{\textbf{InferWiki16k}} & \multicolumn{3}{c|}{\textbf{CoDEx-m-infer}} \\ \cline{2-10} 
  \multicolumn{1}{|l|}{} & MRR    & Hit@1    & Hit@10    & MRR    & Hit@1    & Hit@10  & MRR    & Hit@1    & Hit@10   \\ \hline
  \textbf{TransE}  &  .357  &  .129  &  .709 & .474 &  .214  &  .842  & .366  & .363 & .567  \\
  \textbf{ComplEx} &  .350  &  .218  & .595  & .537  & .377  &  .789  & .252  &  .160  & .430  \\
  \textbf{RotatE}   & .465 &  .297  & .735  &  .629   & .450  & .883 & .352  &  \textbf{.476}  &  .561 \\
  \textbf{ConvE}   &  \textbf{.575} & \underline{.475}  & .747  &  .748  & \underline{.678}  & .868  & .450 & .369  &  .585 \\
  \textbf{TuckER}   &  .573 & .466  & \underline{.755}  & \textbf{.754} & .677 & \underline{.886} & \textbf{.451}  & .365 & \underline{.603}  \\ \hline
  \textbf{AnyBURL}     &  -  & \textbf{.559}  &  \textbf{.783}  & - &  \textbf{.714} & \textbf{.892}  & -  & \underline{.394} & \textbf{.620} \\ \hline
  \end{tabular}
  \caption{Results of Link Prediction. Bold fonts denote the best scores and underlines highlight the second best.}
\label{tab:perf_lp}
\end{table*}

Since most baselines fail in judging unknown as negative, we now investigate them following open-world assumption to see their ability in recognizing unknown triples. Table~\ref{tab:perf_open_tc} shows the macro performance\footnote{Micro performance is only applicable to binary classification, while open-world evaluation is trinary.} on InferWiki16k. We can see that all of the baseline models perform worse than those under the closed-world assumption. On one hand, the trinary classification is intuitively more difficult than binary classification. On the other hand, it is a rather straightforward method to search two decision thresholds --- one between positive and unknown and the other between unknown and negative. This motivates us future works on advanced models to represent KG, which should also be able to detect the limitation and boundaries of given KG. It is a fundamental capacity of inference to respond ``I do not know'', to avoid false negatives in downstream applications.

\begin{figure}[htb]
  \centerline{\includegraphics[width=0.4\textwidth]{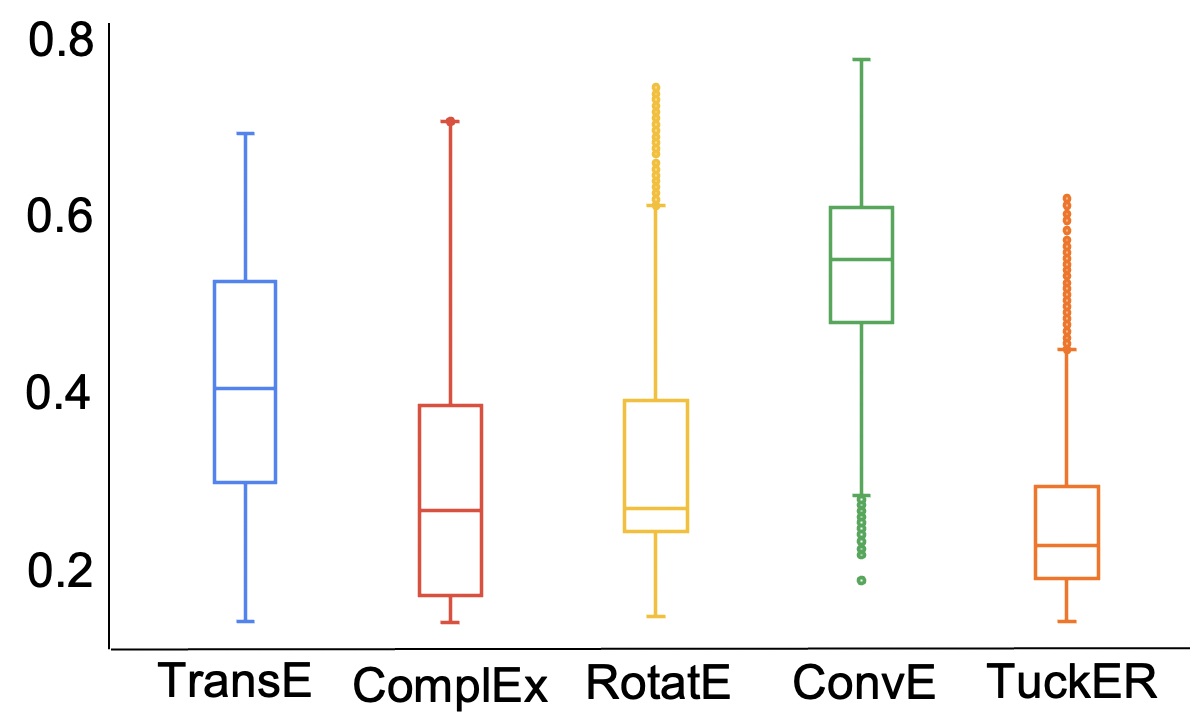}}
    \caption{Macro F1 variance when we search the best thresholds for open-world triple classification.}
  \label{fig:open_thresh}
\end{figure}

Figure~\ref{fig:open_thresh} presents a detailed analysis of each model regarding their search thresholds. We can see that although their best performance seems not bad, the worst scores are only around 10\%. That is, they are very sensitive to thresholds. Besides, most of the time, the average F1 scores of ComplEx, RotatE, and TuckER are around 20\%, while transE achieves higher scores. Maybe that is the reason why it is still the most widely used KGC method. ConvE stably outperforms other baselines, no matter in terms of best, worst, or average performance.

\subsection{Link Prediction Results}
Table~\ref{tab:perf_lp} shows the average scores for head and tail prediction. We can see that (1) AnyBURL performs the best most of the time, but the performance gap is not significant. This is mainly due to its role in dataset construction, although we utilize two sets of triples to minimize rule leakage. Actually, inference of rules may be more important than we thought to improve the reliability and interpretability of knowledge-driven models. This also motivates us to incorporate rule knowledge into KGC training for advanced reasoning ability~\cite{guo2018knowledge,li2019logic}. (2) KGC models perform better on InferWiki16k than InferWiki64k, due to the higher structure density and rule confidence. (3) Models have higher hit@10 and lower hit@1 on InferWiki than other datasets (e.g., CoDEx). This agrees with an intuition that most entities are irrelevant, making it trivial to judge these corrupted triples as in triple classification. And, only a small portion of entities is difficult to predict, which requires strong inference ability. Besides, hit@1 varies a lot, so that we can better compare among models.

\noindent\textbf{Impacts of Inferential Path Length}

\begin{figure}[tb]
    \centerline{\includegraphics[width=0.49\textwidth]{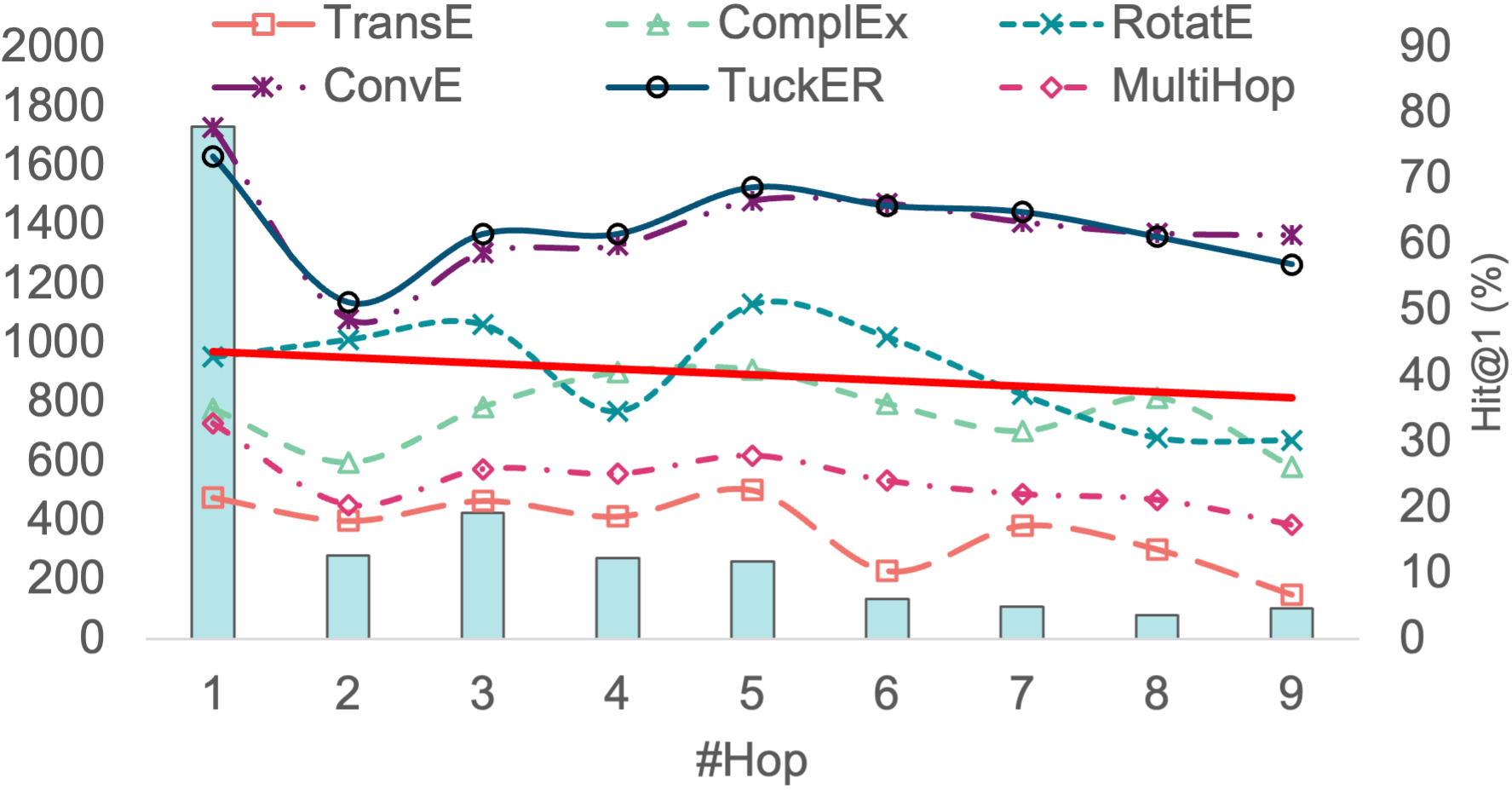}}
      \caption{Hit@1 curves of baseline models for tail prediction. The x-axis denotes the number of hops, and the bars denote the number of examples that have corresponding hops. Red solid line is a performance trend line of six models.}
    \label{fig:length}
    \vspace{-0.3cm}
\end{figure}

Figure~\ref{fig:length} presents Hit@1 curves for tail prediction regarding varying path length on InferWiki64k\footnote{Multihop is designed for tail prediction, and Hit@1 on InferWiki64k is more distinct for following ablation study.}. We can see an overall downwards trend along with the increasing path length. Meanwhile, the large fluctuation may be due to two possible reasons: (1) as discussed in Section~\ref{sec:data_ana}, the inferential paths ensure the predictability, but may not be the shortest ones. This thus offers a conservative analysis and give an upper bound of the performance concerning k-hop paths. Our paths are of high coverage and quality compared with existing datasets, which either conduct case study or post-process datasets via rule mining. (2) Relation types and patterns also have significant impacts. Shorter paths contain more long-tail relations, and longer paths tend to cover many common relations. This improves the difficulty of shorter paths and makes longer paths easier.

\noindent\textbf{Impacts of Relation Patterns}

We present the Hit@1 tail prediction on InferWiki64k regarding relation patterns in Table~\ref{tab:perf_rp}. We can see that symmetry and inversion are not well-solved, which should be considered into evaluation but limited in scale. TransE performs worse on symmetry and inversion relations, consistent with the analysis in~\citet{abboud2020boxe}. Even if ComplEx and RotatE can capture such patterns, they fail to rank corresponding entities at the top. Embedding-based models perform well on hierarchy relations, even outperforms AnyBURL. For compositional relations, it is still quite challenging and worthwhile further investigation.

\begin{table}[]
    \small
    \centering
    \begin{tabular}{r|l|l|l|l|l}
    \hline
    \multicolumn{1}{l|}{} & Sym & Inv & Hier & Comp & Others \\ \hline
    \textbf{TransE}       &  .000  &  .049   &  .479  & .211  & .296 \\
    \textbf{ComplEx}      & .130  & .279 &  .502 & .368  & .414 \\
    \textbf{RotatE}       & .191  & .246 &  .694 & .477  & .610 \\
    \textbf{ConvE}        & .558 & .668 & .855 &  .602   & .784  \\
    \textbf{TuckER}       & .527  & .612 &  .850  &  .625	 & .753  \\
    \textbf{Multihop}     & .231 & .309 & .345  & .240 &  .296 \\ \hline
    \textbf{AnyBURL}      &  .782  &  .793	 &  .782 &  .686 &  .809 \\ \hline
    \end{tabular}
    \caption{Hit@1 tail prediction on Relation Patterns.}
  \label{tab:perf_rp}
\end{table}

\subsection{Comparison of CoDEx-infer and CoDEx}

\begin{figure}[htb]
  \centering
  \subfigure[Triple classification (F1).]{
  \label{fig:EA}\includegraphics[width=0.23\textwidth]{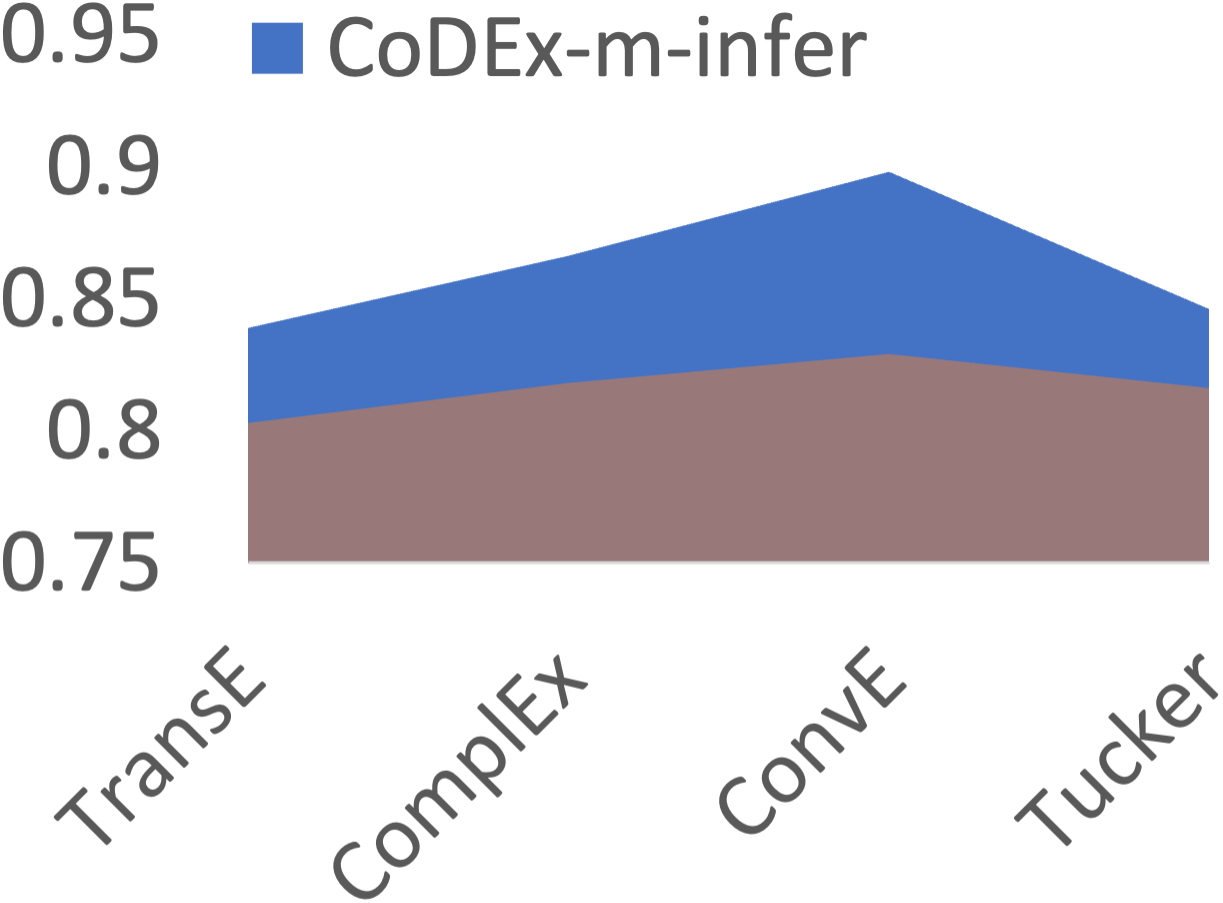}}
  \subfigure[Link Prediction (MRR).]{
  \label{fig:ne_ratio}\includegraphics[width=0.23\textwidth]{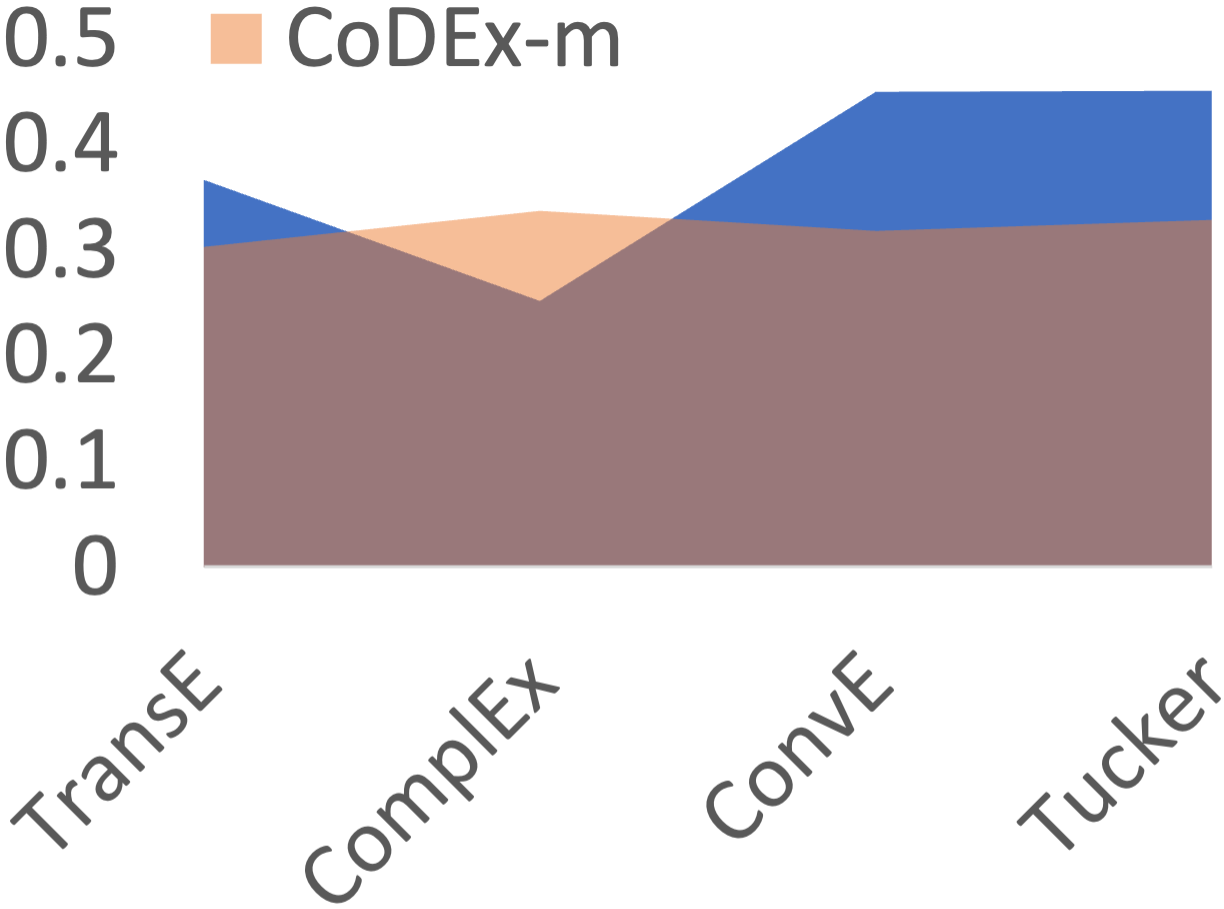}}
  \caption{Comparison of CoDEx-infer and CoDEx.}
  \label{fig:ablation}
\end{figure}

We investigate the impacts of rule-based train/test generatation by comparing CoDEx-m-infer with CoDEx-m. The two datasets share the same training set. The only difference lies in how we obtain the test triples, either using our proposed pipeline (CoDEx-m-infer) or randomly (CoDEx-m). Thus, the results reflect the impacts of inferential guarantee for dataset construction and demonstrate the necessity to avoid over-estimation or under-estimation of the inferential ability of KGC models. We report the performance on CoDEx-m from the original paper~\cite{safavi2020codex}. 

We can see that all of models perform better with inferential path guarantee on CoDEx-m-infer than CoDEx-m, except ComplEx for link prediction. This is because rule guidance elimites those non-inferential testing triples, making the task easier. Nevertheless, the scores on hard cases are actually decreased (as discussed in Figure~\ref{fig:neg} and Table~\ref{tab:perf_rp}). Models are excepted a stronger reasoning ability among several related entities, instead of trivially filtering out massive irrelevant entities. This also demonstrates the necessity of InferWiki to avoid over- or under- estimation of the inferential ability of KGC models --- learning new knowledge from existing ones.

\section{Case Study of Relation Types}
\label{sec:apd_rt}
We illustrate the most frequent relation types and their distribution of InferWiki64k and InferWiki16k in Figure~\ref{fig:app_rel_type}. We can see that InferWiki has a diverse relation types that are not limited to specific domains. Besides, the triples of each relation type are well balanced.

\begin{figure}[tb]
  \centerline{\includegraphics[width=0.49\textwidth]{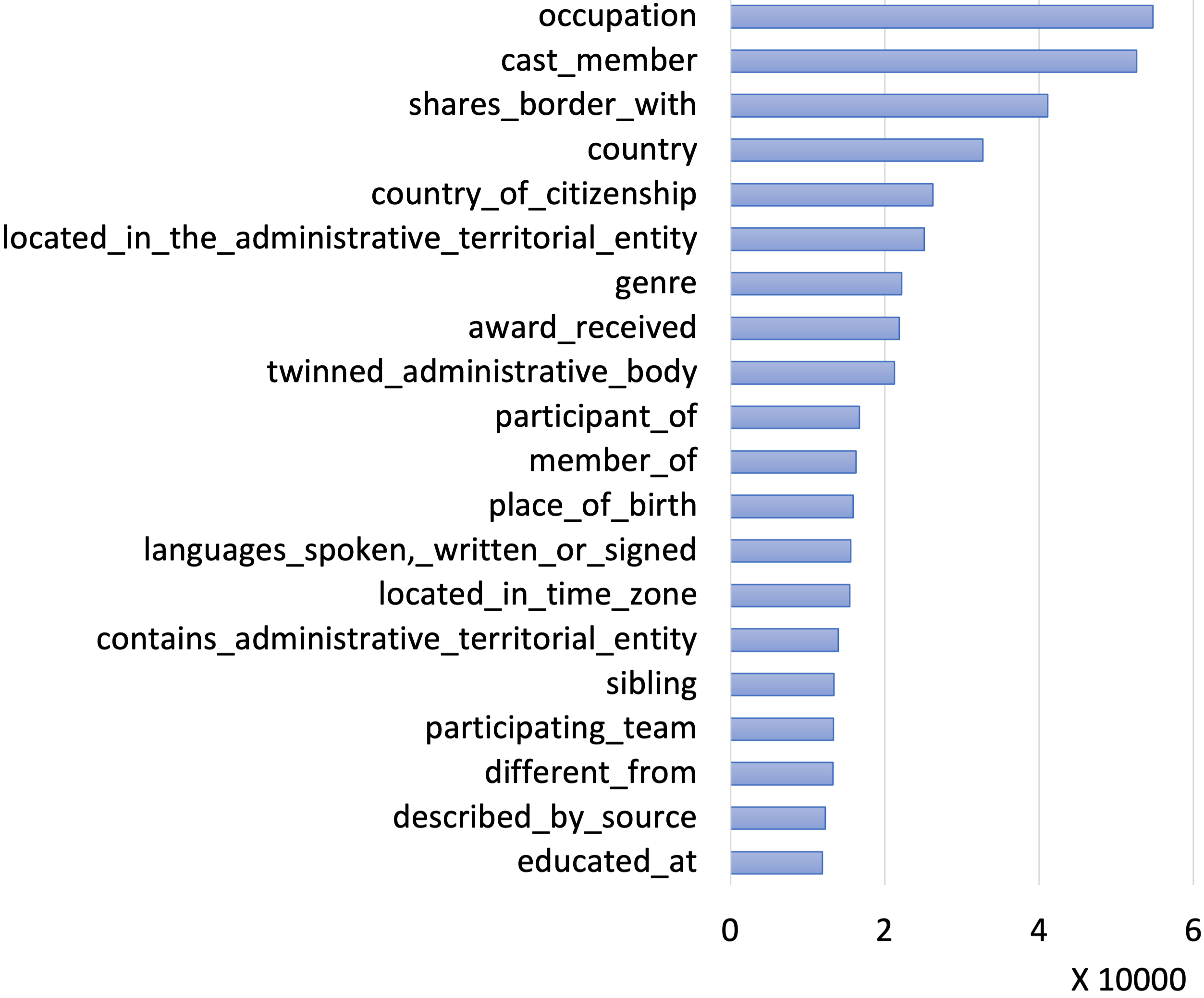}}
    \caption{Distribution of most frequent relation types in InferWiki64k. A comparison with InferWiki16k is in Appendix~\ref{sec:apd_rt}.}
  \label{fig:app_rt64}
\end{figure}

\section{Conclusion}

We highlighted three principles for KGC datasets: inferential ability, assumptions, and patterns, and contribute a large-scale dataset InferWiki. We established a benchmark with three types of seven KGC models on two tasks of triple classification and link prediction. The results present a detailed analysis regarding various inference patterns, which demonstrates the necessity of an inferential guarantee for better evaluation and the difficulty of new open-world triple classification.

In the future, we are interested in cross-KGs inference and transfer~\cite{cao2019multi}, and investigating how to inject knowledge into deep learning architectures, such as for information extraction~\cite{tong2020improving} or text generation~\cite{cao2020expertise}.

\section*{Acknowledgments}
This research was conducted in collaboration with SenseTime. This work is partially supported by A*STAR through the Industry Alignment Fund - Industry Collaboration Projects Grant, by NTU (NTU–ACE2020-01) and Ministry of Education (RG96/20), and by the National Research Foundation, Prime Minister’s Office, Singapore under its Energy Programme (EP Award No. NRF2017EWT-EP003-023) administrated by the Energy Market Authority of Singapore. This work is partially supported by Singapore MOE AcRF T1.

\bibliographystyle{acl_natbib}
\bibliography{ms}

\clearpage
\appendix
\section{Literature Review}
\label{sec:apd_rl}

Table~\ref{tab:app_rl} lists existing KGC datasets. We can roughly classify them into two groups: inferential and non-inferential datasets. The first group are usually manually curated to ensure each testing sample can be inferred from training data through reasoning paths. Families~\cite{garcia2015composing} test family relationships including cousin, ancestor, marriage, parent, sibling, and uncle, among the members of 5 families along 6 generations. Such that there are obvious compositional relationships like uncle $\approx$ sibling + parent or parent $\approx$ married + parent. Kinship~\cite{kemp2006learning} contains kinship relationships among members of the Alyawarra tribe from Central Australia, while Country~\cite{bouchard2015approximate} contains countries, regions, and subregions as entities and is carefully designed to explicitly test the location relationship (i.e., locatedIn and neighbor) among them. The above datasets are clearly limited in scale and inference patterns, thus become not challenging. HOLE~\cite{nickel2016holographic} even achieves 99.7\% ACU-PR on dataset Country~\cite{bouchard2015approximate}.

The second group of datasets are automatically derived from public KGs and randomly split positive triples into train/valid/test, leading to a risk of testing samples non-inferential from training data. FB13~\cite{socher2013reasoning} and FB15K~\cite{bordes2013translating} are commonly used benchmark from FreeBase. FB15k401~\cite{yang2014embedding} is a subset of FB15k containing only frequent relations (relations with at least 100 training examples). To remove test leakage, FB15k-237~\cite{toutanova2015observed} removes all equivalent or inverse relations. Similarly, FB5M~\cite{wang2014knowledge} removes all the entity pairs that appear in the testing set. WN18RR~\cite{dettmers2018convolutional} is the challenging version of WN18~\cite{bordes2013translating} extracted from WordNet. Textual information is also included for specific task, such as FB40K~\cite{lin2015learning} targeting relation extraction dataset New York Times~\cite{riedel}. FB24K~\cite{lin2016knowledge} introduce Attributes. FB15K+~\cite{xie2016representation} introduce types and make fb15k more sparse by only filterring out relation with a frequency lower than one. Another popular knowledge source is YAGO, and the corresponding datasets include YAGO3-10~\cite{dettmers2018convolutional} and YAGO37~\cite{guo2018knowledge}. Except for open-domain KG, NELL~\cite{wang2015knowledge} concentrates on location and sports, and UMLS~\cite{kok2007statistical} targets medical knowledge. 
CoDEx~\cite{safavi2020codex} argues the quality of the above benchmarks, such as NELL995~\cite{xiong2017deeppath} are nonsensical or overly generic. Thus they propose a comprehensive dataset consisting of three knowledge graphs varying in size and structure, entity types, multilingual labels and descriptions, and hard negatives.

\begin{table*}[]
    \small
  \centering
  \begin{tabular}{|l|l|l|l|l|} \hline
  Datasets & source & \#Entity & \#Relation & \#Triples (train/valid/test) \\ \hline
  FB13~\cite{socher2013reasoning} & FreeBase & 75,043 & 13 & 316,232/5,908/23,733 \\ \hline
  FB15k~\cite{bordes2013translating} & FreeBase & 14,951 & 1,345 & 483,142/50,000/59,071 \\ \hline
  FB15k237~\cite{toutanova2015observed} & FreeBase & 14,541 & 237 & 272,115/17,535/20,466 \\ \hline
  FB15k+~\cite{xie2016representation} & FreeBase & 14,951 & 1,855 & 486,446/50,000/62,374 \\ \hline
  FB15k401~\cite{yang2014embedding} & FreeBase & 14,541 & 401 & 560,209/-/- \\ \hline
  FB24k~\cite{lin2016knowledge} & FreeBase & 23,634 & 987 & 402,493/-/21,067 \\ \hline
  FB40k~\cite{lin2015learning} & FreeBase & 39,528 & 1,336 & 370,648/67,946/96,678 \\ \hline
  FB5M~\cite{wang2014knowledge} & FreeBase & 5,385,322 & 1,192 & 19,193,556/50,000/59,071 \\ \hline
  WN11~\cite{socher2013reasoning} & WordNet & 38,696 & 11 & 112,581/2,609/10,544 \\ \hline
  WN18~\cite{bordes2013translating} & WordNet & 40,943 & 18 & 141,442/5,000/5,000 \\ \hline
  WN18RR~\cite{dettmers2018convolutional} & WordNet & 40,943 & 11 & 86,835/3,034/3,134 \\ \hline
  YAGO3-10~\cite{dettmers2018convolutional} & YAGO & 123,182 & 37 & 1,079,040/5,000/5,000 \\ \hline
  YAGO37~\cite{guo2018knowledge} & YAGO & 123,189 & 37 & 989,132/50,000/50,000 \\ \hline
  CoDEx~\cite{safavi2020codex} & Wikidata & 77,951 & 69 & 551,193/30,622/30,622 \\ \hline
  NELL995~\cite{xiong2017deeppath} & NELL & 75,492 & 200 & 154,213/-/- \\ \hline
  NELL$_{loc}$~\cite{wang2015knowledge} & NELL & 672 & 10 & 941/-/- \\ \hline
  Family~\cite{garcia2015composing} & Artificial & 721 & 7 & 8,461/2,820/2,821 \\ \hline
  Kinship~\cite{kemp2006learning} & Artificial & 104 & 26 & 8,548/2,820/2,821 \\ \hline
  Countries~\cite{bouchard2015approximate} & Artificial & 272 & 2 & 1,111/24/24 \\ \hline
  UMLS~\cite{kok2007statistical} & UMLS & 135 & 49 & 5,216/-/- \\ \hline
  \end{tabular}
  \caption{An overview of Knowledge Graph Completion Datasets.}
\label{tab:app_rl}
\end{table*}

\section{Annotation Guideline}
\label{sec:apd_ann}

We provide the following annotation guidelines for annotators to label inferred triples in Section~\ref{sec:labeling}.

\textbf{Task} This is a two-step annotations. First, you must annotate each triple with the label $y\in\{1, -1\}$, where $1$ denotes that the triple is correct and $-1$ denotes that the triple is incorrect. You can find the answer from anywhere you want, such as commonsense, Wikipedia, and professional websites. If you cannot find any evidence to support the statement, you shall choose label $-1$. Second, you must annotate each incorrect triple with the label $\hat{y}\in\{0, -1\}$, where $0$ denotes that you do not know the answer. Now, you can find the answer from our provided triples. If you cannot find any evidence to support the statement, you shall choose label $0$.

\textbf{Examples} Here are some examples judged using three types of knowledge sources.

\begin{itemize}
  \item \textbf{Commonsense}: (Cypriot Fourth Division,	hasPart,	2018–19 Cypriot Third Division) is clearly incorrect, since the fourth division cannot has a part of third division.
  \item \textbf{Professional websites}: To annotate the triple (Bahrain-Merida 2019, hasPart, Carlos Betancur), you may search the person in professional websites, such as \url{https://www.procyclingstats.com/team/bahrain-merida-2019}. Since there is no Carlos Betancur listed in that website, please choose false.
  \item \textbf{Wikipedia}: Given the triples (Tōkaidō Shinkansen,	connectsWith,	Osaka Higashi Line) and (Tōkaidō Shinkansen,	connectsWith,	San’yō Main Line), you can find related station information from the page of Tōkaidō Shinkansen. You can find that Osaka Higashi Line shares a transfer station with Tōkaidō Shinkansen, thus label it with $1$. And, San’yō Main Line doesn't show up in the page, you may label it with $-1$.
\end{itemize}

\section{Relation Patterns}
\label{sec:apd_rp}
InferWiki is able to analyze relation patterns for each path, including symmetry, inversion, hierarchy, and composition, where detailed explanations and examples are listed in Table~\ref{tab:app_rp}.

\begin{table*}[]
    \small
    \centering
    \begin{tabular}{|l|p{3.5cm}|p{10cm}|} \hline
    Pattern & Notation & Example \\ \hline
    symmetry & $r_1(x,y)\Rightarrow r_1(y,x)$ & (Prince Christopher$_{Q44775}$, partner, Friederike$_{Q93614}$) $\Rightarrow$ (Friederike, partner, Prince Christopher) \\ \hline
    inversion & $r_1(x,y)\Leftrightarrow r_2(y,x)$ &  (Amravati district$_{Q1771774}$, capital, Amravati$_{Q269899}$) $\Rightarrow$ (Amravati, capitalOf, Amravati district) \\ \hline
    hierarchy & $r_1(x,y)\Rightarrow r_2(y,x)$ & (Superman$_{Q79015}$, derivativeWork, Superman Returns$_{Q328695}$) $\Rightarrow$ (Superman, presentInWork, Superman Returns) \\ \hline
    composition & $r_1(x,y) \wedge \cdots \wedge r_p(y,z)  \Rightarrow r_{p+1}(x,z)$ & (Eleanor$_{Q156045}$, mother, Joanna$_{Q171136}$) $\wedge$ (Ferdinand I$_{Q150611}$, mother, Joanna) $\wedge$ (Isabella$_{Q157884}$, sibling, Ferdinand I) $\Rightarrow$ (Eleanor, sibling, Isabella) \\ \hline
    \end{tabular}
    \caption{Explanations and examples for various relation patterns. }
\label{tab:app_rp}
\end{table*}

\section{Relation Types}
\label{sec:apd_rt}
\begin{figure*}[htb]
    \centering
    \subfigure[InferWiki64k.]{
    \label{fig:app_rt64}\includegraphics[width=0.48\textwidth]{figure/app_dis_rel64k}}
    \subfigure[InferWiki16k.]{
    \label{fig:app_rt16}\includegraphics[width=0.48\textwidth]{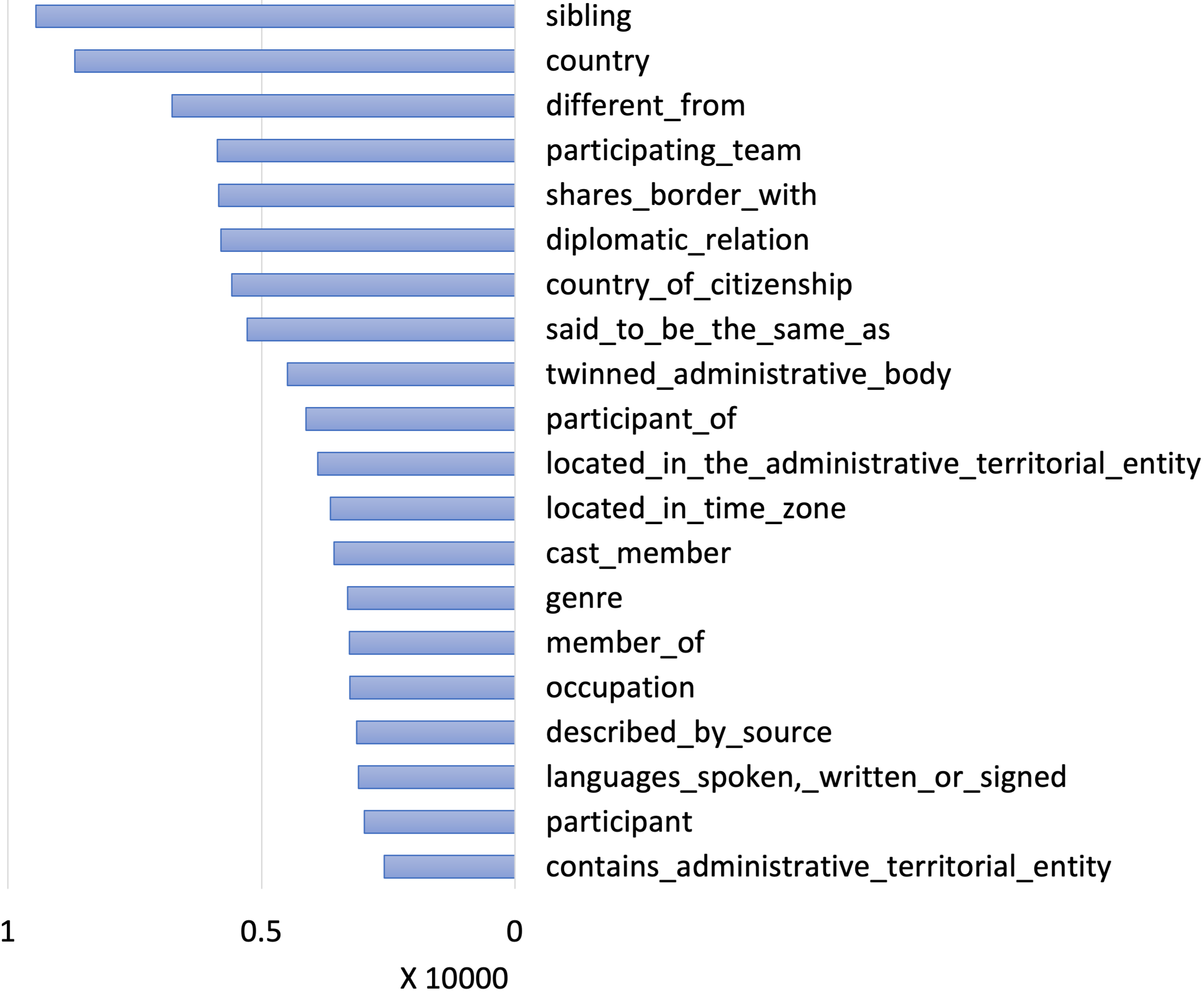}}
    \caption{Distribution of most frequent relation types.}
    \label{fig:app_rel_type}
\end{figure*}

We illustrate the most frequent relation types and their distribution of InferWiki64k and InferWiki16k in Figure~\ref{fig:app_rel_type}.

\section{Comparison with Existing Datasets}
\label{sec:apd_comp}
\begin{figure*}[htb]
    \centering
    \subfigure[Distribution of entities, where x-axis denotes different ranges regarding entity frequency in the train set.]{
    \label{fig:EA}\includegraphics[width=0.98\textwidth]{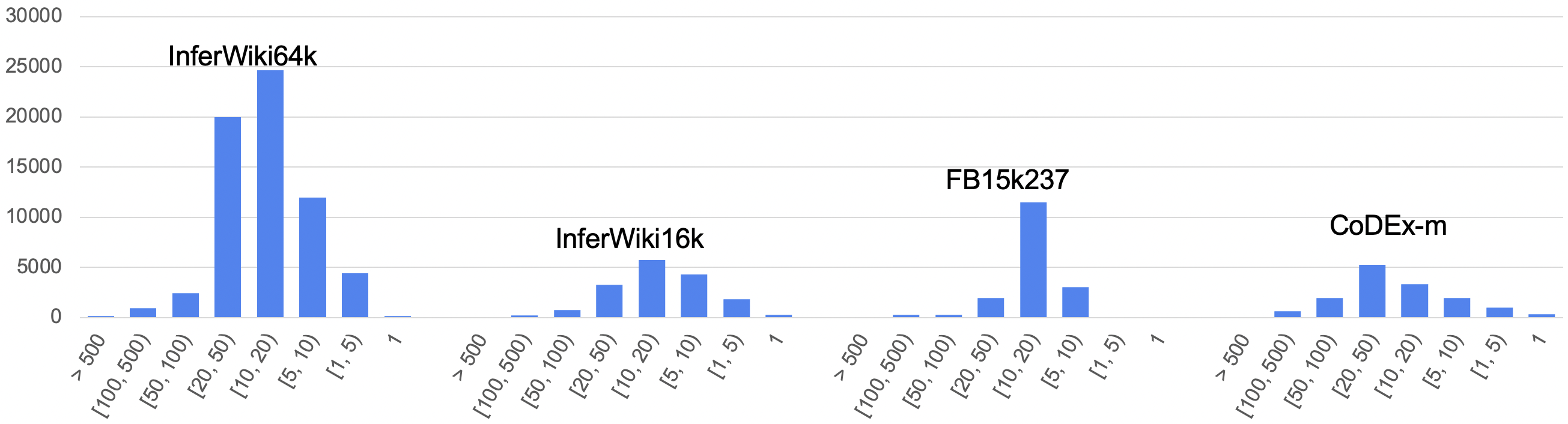}}
    \subfigure[Distribution of entity neighbors, where x-axis denotes ranges regarding average number of neighbors in the train set.]{
    \label{fig:ne_ratio}\includegraphics[width=0.98\textwidth]{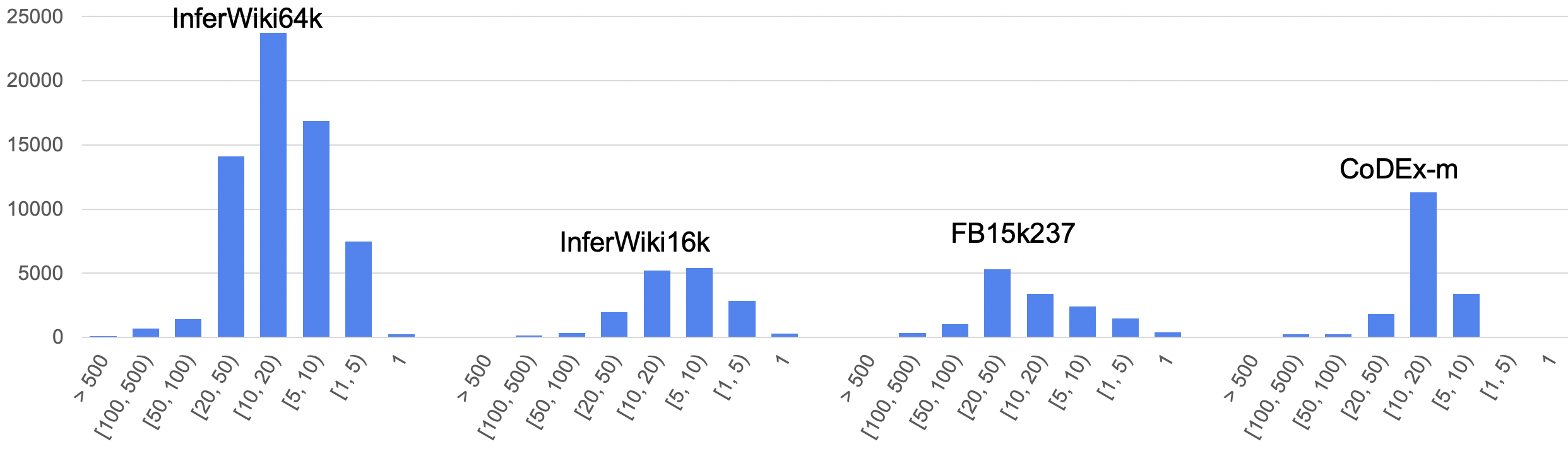}}
    \caption{Distribution of entities and their neighbors.}
    \label{fig:app_ent_dis}
\end{figure*}

Figure~\ref{fig:app_ent_dis} shows the distribution of entities and their neighbors as compared to widely used datasets: FB15k237 and CoDEx-m. 

\section{Experiment Setup}
\label{sec:apd_exp}
Our experiments are run on the server with the following configurations: OS of Ubuntu 16.04.6 LTS, CPU of Intel(R) Xeon(R) CPU E5-2680 v4 @ 2.40GHz, and GPU of GeForce RTX 2080 Ti. We use OpenKE\footnote{https://github.com/thunlp/OpenKE} for re-implementing TransE, ComplEx, and RotatE. For the rest models, we use the original codes for ConvE\footnote{\url{https://github.com/TimDettmers/ConvE}}, TuckER~\footnote{\url{https://github.com/ibalazevic/TuckER}}, Multihop\footnote{\url{https://github.com/salesforce/MultiHopKG}}, and AnyBURL\footnote{\url{http://web.informatik.uni-mannheim.de/AnyBURL/}}. Because we utilize various types of KGC models including embedding-based, multi-hop reasoning (reinforcement learning), and rule-based models, these models largely have their own hyperparameters. To avoid exhaustive parameter search in a large range, we conduct a series of preliminary experiments and find that the suggested parameters work well on Wikidata-based data. We then search the embedding size in the range of $\{256, 512\}$, number of negative samples in the range of $\{15, 25\}$ and margin in the range of $\{4, 8\}$. The optimal parameters of each model on all of three datasets are listed in Table~\ref{tab:para_conf}. The thresholds in triples classification are listed in Table~\ref{tab:app_thred}

\begin{table*}[]
    \small
    \centering
    \begin{tabular}{|l|c|c|c|c|c|c|} \hline
    Hyperparameter & TransE & ComplEx & RotatE & ConvE & TuckER & Multihop \\ \hline
    \multicolumn{7}{|c|}{InferWiki16k} \\ \hline
    Embedding Size & 256 & 512 & 512 & 512 & 512 & 256 \\ \hline
    \# Negatives & 15 & 25 & 25 & - & - & - \\ \hline
    Margin & 4 & 4 & 8 & - & - & - \\ \hline
    Learning Rate & 1.0 & 0.5 & 2e-5 & 1e-4 & 1e-4 & 1e-3 \\ \hline
    Optimizer & SGD & adagrad & adam & adam & adam & - \\ \hline
    Batch Size & 1,625 & 1,625 & 2,000 & 256 & 256 & 128 \\ \hline
    \multicolumn{7}{|c|}{InferWiki64k} \\ \hline
    Embedding Size & 256 & 512 & 512 & 256 & 512 & 256 \\ \hline
    \# Negatives & 15 & 15 & 25 & - & - & - \\ \hline
    Margin & 4 & 4 & 8 & - & - & - \\ \hline
    Learning Rate & 1.0 & 0.5 & 2e-5 & 1e-4 & 1e-4 & 1e-3 \\ \hline
    Optimizer & SGD & adagrad & adam & adam & adam & - \\ \hline
    Batch Size & 7,823 & 7,823 & 2,000 & 256 & 256 & 128 \\ \hline
    \multicolumn{7}{|c|}{CoDEx-m-infer} \\ \hline
    Embedding Size & 512 & 256 & 512 & 256 & 512 & 256 \\ \hline
    \# Negatives & 25 & 25 & 25 & - & - & - \\ \hline
    Margin & 8 & 4 & 4 & - & - & - \\ \hline
    Learning Rate & 1.0 & 0.5 & 2e-5 & 1e-4 & 1e-4 & 1e-3 \\ \hline
    Optimizer & SGD & adagrad & adam & adam & adam & - \\ \hline
    Batch Size & 1,856 & 1,856 & 2000 & 256 & 256 & 128 \\ \hline
    \end{tabular}
    \caption{Best hyperparameter configurations.}
\label{tab:para_conf}
\end{table*}

\begin{table*}[]
    \small\centering
    \begin{tabular}{|c|c|c|c|c|c|c|}
    \hline
    & InferWiki            & TransE                  & ComplEx                 & RotatE                 & ConvE             & TuckER            \\ \hline
    \multirow{4}{*}{\textbf{\begin{tabular}[c]{@{}c@{}}Closed\\ World\end{tabular}}} & \multirow{2}{*}{64k} & {[}-24.4663, -9.0235{]} & {[}-43.0342, 30.6942{]} & {[}-15.7235, 7.8291{]} & {[}0.0, 0.9999{]} & {[}0.0, 0.9982{]} \\ 
    &                      & -16.7449                & -0.2717                 & -0.6498                & 0.1               & 0.01              \\ \cline{2-7} 
    & \multirow{2}{*}{16k} & {[}-24.0588, -4.333{]}  & {[}-21.5906, 24.7742{]} & {[}-21.2362, 7.8282{]} & {[}0.0, 1.0{]}    & {[}0.0, 0.9734{]} \\ 
    &                      & -13.4069                & 2.5191                  & -0.6005                & 0.19              & 0.0097            \\ \hline
    \multirow{2}{*}{\textbf{\begin{tabular}[c]{@{}c@{}}Open\\ World\end{tabular}}}   & \multirow{2}{*}{16k} & {[}-24.0588, -4.333{]}  & {[}-21.5906, 24.7742{]} & {[}-21.2362, 7.8282{]} & {[}0.0, 1.0{]}    & {[}0.0, 0.9734{]} \\ 
    &                      & -16.1685, -11.8288      & -3.5084, 3.4464         & -2.3444, 0.8527        & 0.01, 0.37        & 0.0097, 0.0389    \\ \hline
    \end{tabular}
    \caption{Best thresholds in triple classification, where the upper side is the search range and the lower side is the best values. They are searched on validation.}
\label{tab:app_thred}
\end{table*}

\end{document}